\documentclass{article}
\usepackage[preprint]{neurips_2026}
\usepackage[utf8]{inputenc}
\usepackage[T1]{fontenc}
\usepackage[expansion=false]{microtype}
\usepackage{amsmath,amssymb}
\usepackage{graphicx}
\usepackage{booktabs}
\usepackage{caption}
\usepackage[hidelinks]{hyperref}
\usepackage{xcolor}
\usepackage{pifont}
\usepackage{enumitem}
\usepackage{tikz}
\usetikzlibrary{arrows.meta,positioning}

\newcommand{\eur}{EUR/m$^2$}

\newcommand{\ccby}{CC~BY~4.0}
\newcommand{\setting}[1]{#1}  
\input{generated/numbers.tex}

\title{SlowLab: Benchmarking Autonomous AI Agents on a Year of Experiments They Cannot Undo}
\author{%
  Ningpei Ding\\
  College of Informatics\\
  Huazhong Agricultural University
  \And
  Zai Wen Feng\thanks{Corresponding author.}\\
  College of Informatics\\
  Huazhong Agricultural University\\
  \texttt{zaiwen.feng@mail.hzau.edu.cn}
}

\begin{document}
\maketitle
\suppressfloats[t] 
\begin{abstract}
AI agents are increasingly proposed as autonomous experimenters, yet the benchmarks that test them mostly offer cheap experiments that answer at once. Experiments in physical facilities are different: they take months, occupy scarce equipment and cannot be undone. We present SlowLab, a benchmark in which an agent runs a one-year campaign in four simulated greenhouse compartments on a site whose crop, facility and prices are hidden. The agent decides when to plant which climate policy, whether to stop a crop and which measurements to read while crops grow, and at the end recommends one policy, which a private evaluator scores in weather the agent never saw. In a preregistered study on \res{sites} unseen sites, three low-cost language-model agents ended near a standard grower setting and clearly below a Bayesian-optimisation baseline that starts from that setting. They ran the facility as two synchronous batches, so measurements taken during the season never changed what they planted, and these measurements improved no method's result. Letting other methods re-read the evidence a campaign had collected did not yield significantly better recommendations, which points to where the models start and which experiments they try. Code is released under the MIT licence at \url{https://github.com/dingningpei/slowlab-benchmark}.
\end{abstract}

\section{Introduction}
\label{sec:intro}
\begin{figure}[t]
\centering
\includegraphics[width=\linewidth]{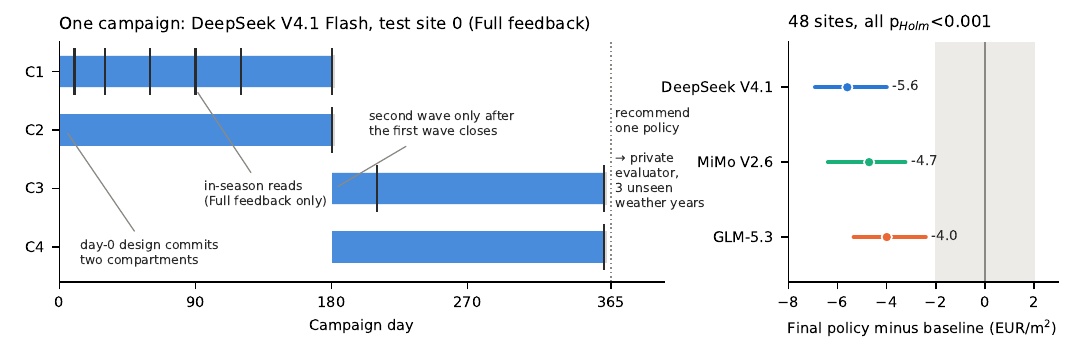}
\caption{SlowLab in one picture. Left: a real campaign year (test site 0, first repeat, Full feedback of the first model alphabetically; selection fixed before plotting). Bars are crops in the four compartments C1--C4 (grey: cleaning), ticks are channel reads; each crop's margin is known only when it closes. Right: across \res{sites} unseen sites, the final policies of three language models scored well below a prior-informed local Bayesian-optimisation baseline (site-level paired differences, BCa 95\% intervals; shaded: the $\pm2$~\eur{} smallest effect of interest).}
\label{fig:teaser}
\end{figure}

Language-model agents are increasingly proposed as autonomous experimenters, and the benchmarks that evaluate them mostly let an agent query an experiment cheaply and get the answer at once \citep{boxinggym,scigym}. Experiments in physical facilities work differently. A greenhouse trial occupies a compartment for half a year, cannot be revised once it has started (a crop can only be continued or destroyed), and reveals its outcome only at harvest, while partial process measurements accumulate in the meantime. An agent in such a facility must decide what to try when every trial is a long, scarce commitment, how to schedule a few units in parallel, and whether to act on the noisy signals that arrive before the outcome. No existing benchmark tests this combination (Table~\ref{tab:related}).

We introduce \textbf{SlowLab}, a benchmark for agents running slow, irreversible experiments (Figure~\ref{fig:teaser}). An agent system, a language model in a harness or a numerical optimiser, runs four simulated greenhouse compartments for one year on a site with hidden cultivar, facility and prices: it decides when to plant which climate policy, whether to stop crops and what to read while they grow, and finally recommends one policy, which a private evaluator scores in three unseen weather years. Under Full feedback the agent can read every process record as it arrives, under Endpoint feedback only the totals of each finished crop.

Our contributions are:
\begin{itemize}[leftmargin=*,itemsep=1pt,topsep=2pt]
\item \textbf{The benchmark.} An environment whose experiments are slow, irreversible, bound to four shared compartments and observed through delayed, partial evidence, a combination no existing benchmark offers (Table~\ref{tab:related}), with an audited agent interface and validity checks showing that the score separates strategies and that runs replay bit for bit on other hosts (Section~\ref{sec:benchmark}).
\item \textbf{Diagnostic experiments.} Beyond one score, two preregistered designs locate where an agent fails: branched campaigns that share the day-0 design isolate the value of in-season measurements, and a Reader swap, in which other methods re-read a campaign's final evidence, separates gathering evidence from reading it (Section~\ref{sec:setup}).
\item \textbf{Three findings} on \res{sites} unseen sites with three low-cost language models (Section~\ref{sec:results}): \textbf{(i)} the models end near a standard grower setting, \res{range.E1.abs.lo}--\res{range.E1.abs.hi}~\eur{} below the baseline; \textbf{(ii)} no method, the baseline included, profits from in-season process feedback, and the models run the facility as two synchronous batches, so in-season readings never change what they plant; \textbf{(iii)} the final reading of the evidence is not the bottleneck, which points to how the agents start and explore.
\end{itemize}

\section{Related work}
\label{sec:related}
\begin{table}[t]
\centering\small
\caption{SlowLab against the closest benchmarks and environments. Columns: outcomes arrive only after a long delay during which the agent keeps acting; starting an experiment commits a resource that cannot be released without loss; a small fixed set of units must be scheduled in parallel; partial measurements are readable while an experiment runs; each instance has hidden parameters to be learned by experimenting; the final output is scored under conditions the agent never observed; a non-language-model baseline is evaluated alongside language-model agents; the evaluation was preregistered. \ding{51}: yes; $\sim$: partly; --: no; n/a: no language-model agents; ?: not reported.}
\label{tab:related}
\resizebox{\linewidth}{!}{%
\begin{tabular}{@{}lcccccccc@{}}
\toprule
 & Delayed & Irrevers- & Parallel & Mid-run & Hidden & Unseen-con- & Non-LLM & Pre- \\
 & outcomes & ible & units & signals & instance & dition score & baseline & registered \\
\midrule
BoxingGym \citep{boxinggym} & -- & -- & -- & -- & \ding{51} & $\sim$ & $\sim$ & -- \\
SciGym \citep{scigym} & -- & -- & -- & -- & \ding{51} & \ding{51} & -- & -- \\
CARE \citep{care2026} & -- & -- & -- & -- & $\sim$ & -- & \ding{51} & $\sim$ \\
LLMs for BO \citep{llmbo2025} & -- & -- & $\sim$ & -- & $\sim$ & -- & \ding{51} & -- \\
Vending-Bench \citep{vendingbench} & $\sim$ & $\sim$ & $\sim$ & \ding{51} & \ding{51} & -- & $\sim$ & -- \\
GreenLight-Gym \citep{vanlaatum2025greenlightgym} & -- & -- & -- & \ding{51} & $\sim$ & -- & n/a & -- \\
WOFOSTGym \citep{wofostgym} & \ding{51} & $\sim$ & $\sim$ & \ding{51} & $\sim$ & ? & n/a & -- \\
\midrule
SlowLab (ours) & \ding{51} & \ding{51} & \ding{51} & \ding{51} & \ding{51} & \ding{51} & \ding{51} & \ding{51} \\
\bottomrule
\end{tabular}}
\end{table}

\paragraph{Benchmarks of experimenting agents.} BoxingGym gives language agents up to ten sequential experiments in each of ten generative probabilistic environments and scores experimental design by expected information gain and model discovery by prediction error \citep{boxinggym}. SciGym asks agents to recover the missing reactions of systems-biology models by running simulated experiments in a dry lab, and scores the recovered model under perturbed conditions \citep{scigym}. In both, every experiment returns its outcome at once and costs nothing but a query (Table~\ref{tab:related}). Vending-Bench tests long-horizon coherence in a simulated business with delayed deliveries and hidden demand parameters, scored by the net worth at the end of the same run \citep{vendingbench}. SlowLab combines what these settings lack: experiments that occupy one of four units for half a year, cannot be revised, and are judged by a policy recommendation scored in weather the agent never saw.

\paragraph{Language models against classical optimisers.} CARE puts a language model inside a budgeted sequential reaction-optimisation loop: the model writes an executable scoring program, and a gate conditioned on a non-LLM reference policy decides whether the program's candidate replaces the reference choice; evaluation replays recorded campaigns offline \citep{care2026}. A study of language models for Bayesian optimisation in gene-perturbation and molecular design found them insensitive to experimental feedback and below classical baselines \citep{llmbo2025}. Reaction and materials benchmarks provide noisy executable objectives and planner libraries \citep{summit,olympus,minerva}; budgeted, delayed-feedback and trust-region Bayesian optimisation address evaluation budgets, pending evaluations and local search \citep{astudillo2021budgeted,verma2022delayed,eriksson2019turbo}, and our baseline uses the last with a prior taken from a standard grower setting.

\paragraph{Crop simulators and scientific agents.} Crop-simulator environments expose crop models as control problems for reinforcement learning: WOFOSTGym for field crops \citep{wofostgym} and GreenLight-Gym, built on the GreenLight greenhouse model \citep{katzin2020greenlight}, for greenhouse climate control under parametric uncertainty \citep{vanlaatum2025greenlightgym}. SlowLab uses the same simulator, but evaluates how an agent designs, monitors and reads a year of irreversible experiments rather than how it controls the climate step by step. Scientific-agent suites mostly start from existing data, and autonomous laboratories connect planners to instruments \citep{scienceagentbench,discoverybench,coscientist,chemcrow,szymanski2023autonomous,burger2020mobile}.

\section{The SlowLab benchmark}
\label{sec:benchmark}
\subsection{Design principles}
\label{sec:principles}
SlowLab is designed around what makes experiments in physical facilities hard, and each property is a requirement on the environment. Experiments are \textbf{Slow}: a crop occupies a compartment for 180 days, so a 365-day campaign leaves room for about two waves. They are \textbf{Irreversible}: starting a crop commits a compartment, and stopping it discards the crop without salvage. They are \textbf{Facility-bound}: four compartments must be shared, in parallel and on schedules the agent chooses. Their evidence is \textbf{Delayed and partial}: a crop's margin is known only when it closes, and while it grows only noisy process measurements exist. The site is \textbf{Hidden}: cultivar, facility and prices are drawn per site and never shown, so the agent must learn the site from its own experiments. Scoring is \textbf{Private and out of sample}: the recommended policy is scored in weather the agent never saw. No benchmark in Table~\ref{tab:related} combines the first four; the last two keep the score from rewarding memorisation of the site or of the campaign weather.

\begin{figure}[t]
\centering
\includegraphics[width=\linewidth]{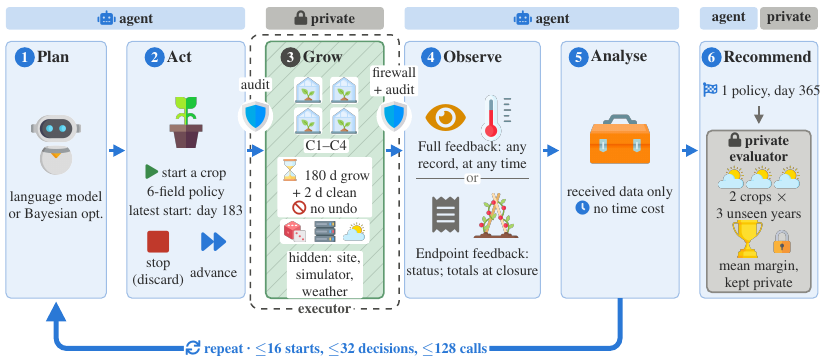}
\caption{One campaign through the agent's eyes; Section~\ref{sec:steps} describes steps 1--6. Everything inside the dashed executor is hidden from the agent. Shields: every message is hash-audited, and messages to the model also pass the information firewall. The language-model harness is shown in Appendix Figure~\ref{fig:harness}. Icons: Freepik, flaticon.com.}
\label{fig:framework}
\end{figure}
\subsection{Problem formulation}
\label{sec:formal}
A site $s$ draws hidden parameters $\theta_s$ (cultivar, facility and prices; Section~\ref{sec:steps}) and a campaign weather year. An agent system, a language model in a harness or a numerical optimiser, runs a facility of $K=4$ compartments for $T=365$ days. A policy $\pi\in\Pi$ fixes six set-points for a whole crop: day and night temperature, CO$_2$ target, supplemental light hours, ventilation temperature offset and ventilation humidity threshold, inside a box with night temperature not above day temperature. An experiment $(k,t_0,\pi)$ grows a crop under $\pi$ in compartment $k$ from day $t_0\le 183$ for 180 days, followed by 2 days of cleaning; a stop at $t<t_0+180$ discards it. Let $m(\pi,d;\theta_s,w)$ be the contribution margin per m$^2$ of a crop grown under $\pi$ from planting date $d$ on site $s$ in weather year $w$: harvest revenue minus the cost of electricity, heat, CO$_2$, planting, cleaning and daily overhead, at the site's public prices. An experiment's margin is released only when it closes. After day $T$ the agent recommends one policy $\hat\pi$, scored as
\[
J_s(\hat\pi)=\frac{1}{6}\sum_{w\in W_s}\ \sum_{d\in\{\text{1 Jan},\,\text{2 Jul}\}} m\!\left(\hat\pi,d;\theta_s,w\right),
\]
where $W_s$ holds three evaluation weather years not used in the campaign. The agent knows this rule from the task description, but the evaluator is private and the score is never shown to it. The benchmark score is the mean of $J_s$ over unseen sites. A campaign allows at most 16 experiments, 32 decision calls (start, stop, recommend) and 128 interface calls; a language-model agent may also send at most $5\times10^6$ prompt characters.

\subsection{A campaign step by step}
\label{sec:steps}
\label{sec:env}
\label{sec:interface}
\paragraph{Plan and act (steps 1--2).} The agent acts only through the executor, with five actions: start an experiment $(k,t_0,\pi)$, stop a running crop, advance the clock, observe a channel and recommend $\hat\pi$. For language models a harness turns planning into calls: each carries the task, the tool catalogue, a digest of the state, the six most recent turns in full and a 2{,}000-character notebook the model maintains, and the model answers with one JSON object (Appendix~\ref{app:interface}). Malformed replies and invalid actions count as the method's own result; when the budget is nearly spent, the harness advances to day $T$ and accepts only a recommendation, falling back to the pre-experiment recommendation.

\paragraph{Grow (step 3).} The compartments are 96~m$^2$ greenhouses, the size of research compartments used in tomato trials, simulated with the GreenLight model \citep{katzin2020greenlight} under historical outdoor weather from the Cabauw observatory \citep{knmi_cabauw,knmi_cabauw_rad} (Appendix Table~\ref{tab:task}). The site parameters $\theta_s$ hold cultivar parameters (temperature window, photosynthetic capacity, fruit growth, earliness, maximum LAI), facility parameters (cover transmission, leakage, ventilation, side-wall heat exchange, LED efficacy, deep-soil temperature), prices and small per-compartment differences. Their distribution was fixed from physical and economic plausibility before any method was run and is never filtered on outcomes. Each site uses one campaign year and the three years of $W_s$, drawn without replacement from eleven audited years (2002--2011, 2015); development work used separate years and a separate master seed.

\paragraph{Observe (step 4).} While a crop runs, eight channels (climate, a canopy proxy, cumulative harvest and resource use) are measured every 300~s. Under \emph{Full} feedback any record measured so far can be read at any time; under \emph{Endpoint} feedback the agent sees only each compartment's status and each crop's totals at closure. Because a language model's replies are sampled, each of its campaigns is \emph{branched}: the day-0 design is made once without knowing the condition, then a Full and an Endpoint branch continue from it, differing only in feedback. An information firewall rejects any outgoing message that would reveal hidden site information, and an audit stores every message with its hash, so every campaign can be replayed.

\paragraph{Analyse (step 5).} Analysis tools summarise the agent's own data: a crop table, channel summaries, a Gaussian-process fit of margin on $(\pi,t_0)$ with predictions, a process predictor of a running crop's final margin trained on development sites, and a space-filling candidate generator over $\Pi$. They do not advance time.

\paragraph{Recommend (step 6).} After day $T$ the agent submits $\hat\pi$, and the private evaluator computes $J_s(\hat\pi)$ by simulating the two scoring crops in each year of $W_s$.

\subsection{Reference points}
\label{sec:refs}
A score in \eur{} means little on its own, so two references that use no agent bracket it. The \emph{fixed reference} applies a standard grower setting, chosen on development sites, without experiments; beating it shows that a year of experiments paid off. The \emph{best-known reference} is the best policy found by a numerical search on a random subset of test sites; it shows how much room is left above the fixed reference and is a lower bound on the attainable score.

\subsection{Validity of the benchmark}
\label{sec:validity}
\paragraph{The score separates strategies.} On the test sites the fixed reference scored \res{mean.fixed.reference}~\eur{}, the main baseline (Section~\ref{sec:setup}) \res{mean.baseline.full}, and the best-known policies lay \res{ref.headroom}~\eur{} above the fixed reference on the searched sites, so the scale has room both above and below a competent method. With three evaluation years, the per-site standard error of a paired difference between two policies is \res{valid.evalerr.paired}~\eur{} on development sites, well below the 2~\eur{} smallest effect of interest; with \res{sites} sites the smallest differences detectable with 80\% power ranged from \res{range.mde.lo} to \res{range.mde.hi}~\eur{} across the primary tests.
\paragraph{Runs are reproducible.} The simulator, the evaluator and the harness are deterministic given the recorded model replies. \res{valid.trace.campaigns} formal campaigns replayed on a different host produced \res{valid.trace.identical} traces and settlements, and \res{valid.replay.evals} formal evaluations rerun after the study were \res{valid.replay.identical}.

\section{Evaluation setup}
\label{sec:setup}
\paragraph{Language-model agents.} DeepSeek V4.1 Flash (official API, thinking disabled), MiMo V2.6 Flash (Xiaomi endpoint) and GLM-5.3 Flash (Z.AI endpoint; reasoning mandatory there and capped at 1{,}024 tokens, a recorded asymmetry), each through its maker's endpoint, temperature 0.7, at most 2{,}048 output tokens per call, in the harness of Section~\ref{sec:interface}. Each model first gives a pre-experiment recommendation from the public task only. Its campaigns are branched as in Section~\ref{sec:steps}.

\paragraph{Main baseline: prior-informed local BO.} The numerical baseline uses the same interface, budget and tools. It starts from the fixed reference in one compartment and orthogonal-design perturbations around it in another, adds two staggered compartments at mid-season (under Full feedback chosen with the public process predictor), and plants a second wave when the first crops finish. Its Gaussian process has a season-interaction kernel whose hyperparameters were fitted once on development data and frozen; expected improvement is maximised only inside a box trust region around the best tried policy \citep[TuRBO rules,][]{eriksson2019turbo}, with an initial radius of \res{radius.selected}\% of each range (the largest of three candidates) chosen on development sites by a rule fixed in advance; only tried policies are recommended. This baseline replaced a plain BO after the pilot showed the plain BO scoring below the fixed reference (Appendix~\ref{app:baseline}); the decision referred only to the fixed reference, never to language-model results.

\paragraph{Knowledge asymmetry.} The comparison is not knowledge-symmetric: the baseline starts from the fixed reference and uses a kernel fitted on development data; the models are given neither, and start from their own pre-experiment recommendations. E1 therefore compares general-purpose agents with a domain-informed optimiser.

\paragraph{Research questions.} Three preregistered questions, each answered by one experiment:
\textbf{RQ1} (E1) How good is the final policy a model finds, relative to the baseline with the same interface and budget?
\textbf{RQ2} (E2) Does reading process measurements during the season (Full) improve on seeing only end-of-crop totals (Endpoint)?
\textbf{RQ3} (E3) Are poor recommendations caused by the evidence gathered or by how it is read? A \emph{Reader} here is step 6 of Figure~\ref{fig:framework} taken on its own: a procedure that turns a campaign's final evidence into one recommended policy.

\paragraph{Design and statistics.} \res{sites} unseen test sites; each language model ran two repeats per site, each a branched Full/Endpoint campaign, and the baseline ran two seeds under each condition. The site is the unit: repeats are averaged within a site, and the sample size came from a power analysis on eight pilot sites. The primary family has seven tests with Holm correction: E1, each model's Full final policy minus the baseline's, and E2, Full minus Endpoint for each of the four methods. Each is a site-level paired mean difference, tested by a two-sided sign-flip permutation test ($10^5$ random sign flips) and given a bias-corrected and accelerated (BCa) bootstrap 95\% interval ($10^4$ site resamples); differences below the smallest effect of interest, 2~\eur{} fixed in advance, count as practically negligible. Secondary results, with Benjamini--Hochberg correction over one joint family, are each model's gain over its pre-experiment recommendation, each method against the fixed reference, the between-method differences of feedback gains and E3. Format errors, budget exhaustion and fallback recommendations are the method's own result.

\paragraph{Preregistration and deviations.} Sample size, tests, models, prompts, code and configuration were locked by hash before the run, and the analysis code before any score existed (Appendix~\ref{app:protocol}). Campaigns ran in site order rather than random order; all methods of a site ran in the same window. After the first part of the run, a field-order error in the inputs of the process predictor and the baseline was found; under a new lock, the affected records (all baseline campaigns and \res{repair.branches} language-model Full branches) were rerun from the point of the error, with the baseline radius re-selected by the preregistered rule. Results use the corrected data; the original results are in Appendix~\ref{app:original}.

\section{Results}
\label{sec:results}
\begin{table}[t]
\centering\small
\caption{Mean score of the recommended policy over the \res{sites} test sites (\eur{}, three evaluation years each). The fixed reference runs no experiments, so its score is the same in every column.}
\label{tab:means}
\begin{tabular}{lccc}
\toprule
Method & Full & Endpoint & Pre-experiment \\
\midrule
Prior-informed local BO (baseline) & \res{mean.baseline.full} & \res{mean.baseline.endpoint} & -- \\
GLM-5.3 Flash & \res{mean.glm.full} & \res{mean.glm.endpoint} & \res{mean.glm.initial} \\
MiMo V2.6 Flash & \res{mean.mimo.full} & \res{mean.mimo.endpoint} & \res{mean.mimo.initial} \\
DeepSeek V4.1 Flash & \res{mean.deepseek.full} & \res{mean.deepseek.endpoint} & \res{mean.deepseek.initial} \\
Fixed reference (no experiments) & \res{mean.fixed.reference} & \res{mean.fixed.reference} & \res{mean.fixed.reference} \\
\bottomrule
\end{tabular}
\end{table}

\begin{figure}[t]
\centering
\includegraphics[width=\linewidth]{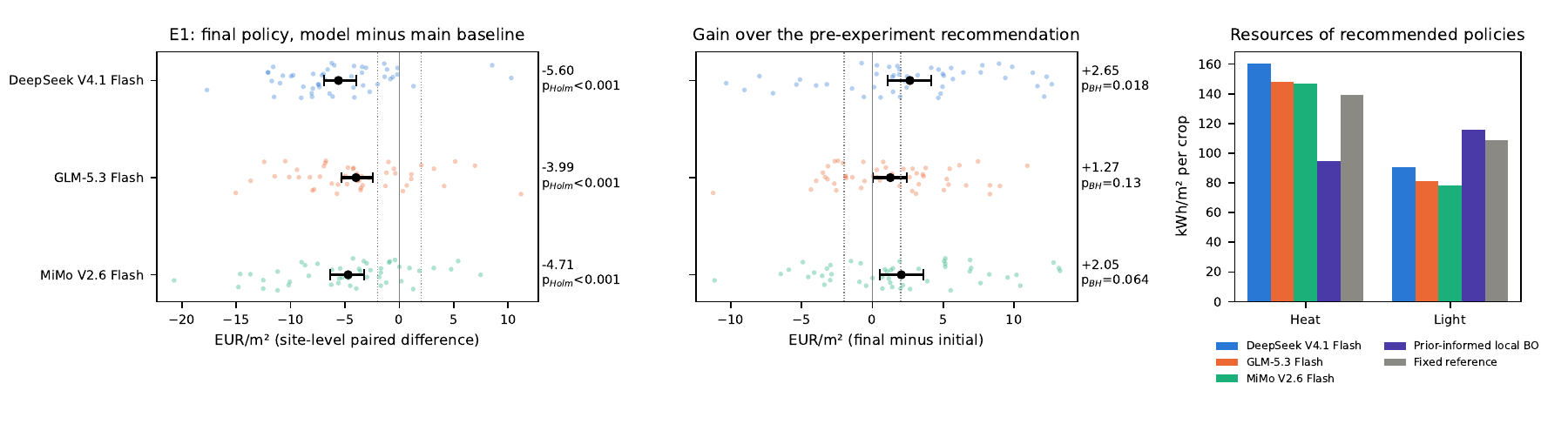}
\caption{Language models end near the grower setting, well below the baseline. Left: site-level paired difference between each model's final (Full) policy and the baseline's; dots are sites, bars BCa 95\% intervals, dotted lines the smallest effect of interest ($\pm2$~\eur{}); p-values Holm-adjusted. Middle: gain over the model's own pre-experiment recommendation (BH-adjusted). Right: heat and light per crop of the recommended policies in the evaluation runs.}
\label{fig:e1}
\end{figure}

\subsection{Language models end near the grower setting, well below a domain-informed optimiser (RQ1, E1)}
\label{sec:rq1}
All three models ended clearly below the baseline (Figure~\ref{fig:e1}, Table~\ref{tab:means}): DeepSeek by \res{E1.deepseek.mean}~\eur{} [\res{E1.deepseek.lo}, \res{E1.deepseek.hi}], MiMo by \res{E1.mimo.mean} [\res{E1.mimo.lo}, \res{E1.mimo.hi}] and GLM by \res{E1.glm.mean} [\res{E1.glm.lo}, \res{E1.glm.hi}], each with $p_\text{Holm}$ \res{E1.glm.p}; every interval lies beyond the smallest effect of interest. The models did end better than they started, by \res{learn.deepseek.mean} (DeepSeek, $p_\text{BH}$ = \res{learn.deepseek.p}), \res{learn.mimo.mean} (MiMo, \res{learn.mimo.p}) and \res{learn.glm.mean}~\eur{} (GLM, \res{learn.glm.p}), significantly only for DeepSeek, and stopped near the fixed reference (\res{vsfixed.glm.mean}, \res{vsfixed.mimo.mean} and \res{vsfixed.deepseek.mean}~\eur{}), which the baseline exceeded by \res{vsfixed.baseline.mean} [\res{vsfixed.baseline.lo}, \res{vsfixed.baseline.hi}].

The gap has two parts. The first is the starting point: the baseline starts from the fixed reference (\res{mean.fixed.reference}~\eur{}), which the models are not given, while their own pre-experiment recommendations scored \res{range.initial.lo}--\res{range.initial.hi}. The second is the search itself: within the year the baseline gained \res{vsfixed.baseline.mean}~\eur{} over its starting point and the models \res{range.learn.lo} to \res{range.learn.hi} over theirs. The two kinds of policy also differ in style: the models' final policies used less supplemental light than the baseline's (\res{range.llm.full.light.lo}--\res{range.llm.full.light.hi} against \res{res.baseline.full.light}~kWh/m$^2$ per crop) and more heat (\res{range.llm.full.heat.lo}--\res{range.llm.full.heat.hi} against \res{res.baseline.full.heat}).

\begin{figure}[t]
\centering
\includegraphics[width=0.75\linewidth]{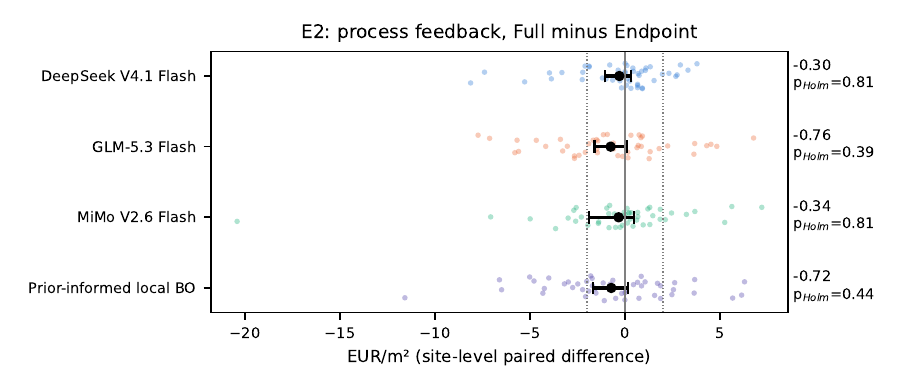}
\caption{No method profits from in-season process feedback: site-level paired difference between the Full and Endpoint branches of each method (Holm-adjusted).}
\label{fig:e2}
\end{figure}

\subsection{No method profits from in-season process feedback (RQ2, E2)}
\label{sec:rq2}
Being able to read process measurements during the season did not improve the final policy of any method (Figure~\ref{fig:e2}). Full minus Endpoint was \res{E2.deepseek.mean} [\res{E2.deepseek.lo}, \res{E2.deepseek.hi}] for DeepSeek, \res{E2.glm.mean} [\res{E2.glm.lo}, \res{E2.glm.hi}] for GLM, \res{E2.mimo.mean} [\res{E2.mimo.lo}, \res{E2.mimo.hi}] for MiMo and \res{E2.baseline.mean} [\res{E2.baseline.lo}, \res{E2.baseline.hi}]~\eur{} for the baseline; none is significant, every upper bound lies well below the 2~\eur{} effect of interest, and the feedback gains did not differ between methods (all $p_\text{BH}\ge$ \res{range.gaindiff.p.lo}).

The campaign records show why: the models ran the facility as two synchronous batches. In \res{range.fb.waitpct.lo}--\res{range.fb.waitpct.hi}\% of their Full branches the first new crop was planted only after the first wave had closed and released its totals, so in-season readings could not change what was planted next (Appendix Figure~\ref{fig:timelines}). The design did not force this: crops may start up to day 183, so idle compartments could have been planted at mid-season on readings of the first wave, as the baseline does. Before that planting the Full branches read a sensor channel only \res{range.fb.reads.lo}--\res{range.fb.reads.hi} times on average and called the process predictor \res{range.fb.predictor.lo}--\res{range.fb.predictor.hi} times. A running crop was stopped in a single campaign (a GLM Full branch, \res{fb.glm.stops} of \res{fb.glm.campaigns}), and the Full and Endpoint branches planted their later crops in the same compartments on the same days in \res{fb.deepseek.sameschedule}, \res{fb.glm.sameschedule} and \res{fb.mimo.sameschedule} of \res{fb.deepseek.campaigns} campaigns (DeepSeek, GLM, MiMo). The models' stated reasons on test site 0 (first repeat, the rule used for Figure~\ref{fig:teaser}) are typical: \emph{``Let the baseline crops finish their 180-day cycle so totals are released''} (DeepSeek, day 150) and, after consulting the process predictor at day 30, \emph{``Let the four batch-1 crops run to maturity; results will release at closure\ldots''} (GLM). The baseline, which by construction staggers two compartments at mid-season with the process predictor, did not gain from the feedback either. Both branches receive every finished crop's margin, good or bad, and these totals seem to carry most of what the next planting needs, so the in-season signal available through the predictor added little in this task.

\begin{figure}[t]
\centering
\includegraphics[width=0.85\linewidth]{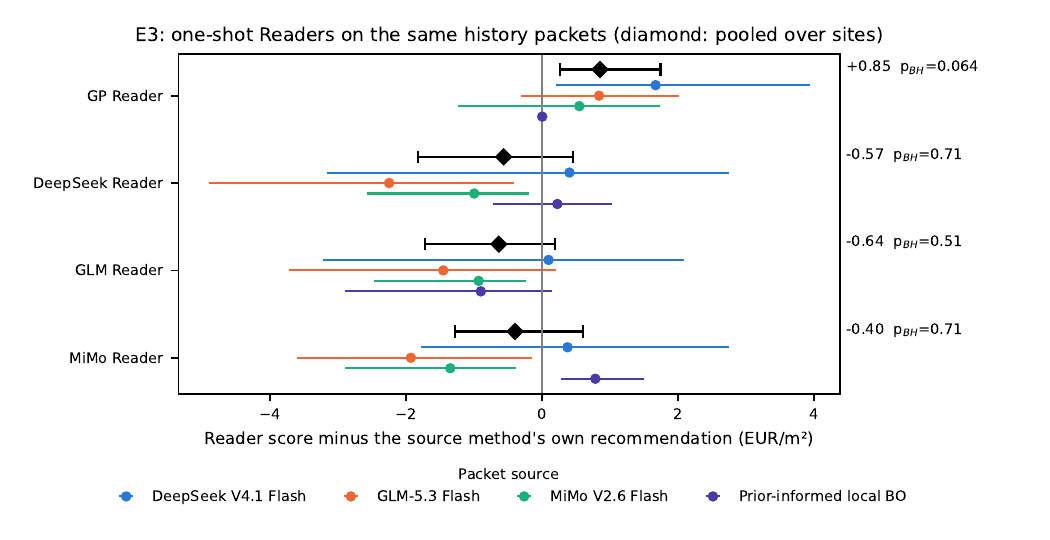}
\caption{Reading the final evidence is not the bottleneck. Reader swap gain: each Reader's recommendation from a history packet minus the recommendation of the method that collected the packet, by packet source (16 packets each, BCa 95\% intervals) and pooled over sites (diamonds; BH-adjusted over the joint secondary family).}
\label{fig:e3}
\end{figure}

\subsection{Reading the final evidence is not the bottleneck (RQ3, E3)}
\label{sec:rq3}
From the Full branches of each method's first repeat, a committed seed drew 64 history packets, 16 per method on distinct sites; a packet holds the public evidence the method had before its final recommendation and names no method. Each packet was read by the GP Reader (the baseline's own recommendation step) and by each model in one shot. The \emph{Reader swap gain} is the Reader's recommendation score minus the source method's own on the same packet (BH over the joint secondary family of \res{secondary.family} tests).

No Reader's pooled swap gain differed significantly from zero (Figure~\ref{fig:e3}): \res{E3.deepseek.mean}~\eur{} [\res{E3.deepseek.lo}, \res{E3.deepseek.hi}] for DeepSeek ($p_\text{BH}$ = \res{E3.deepseek.p}), \res{E3.glm.mean} [\res{E3.glm.lo}, \res{E3.glm.hi}] for GLM (\res{E3.glm.p}), \res{E3.mimo.mean} [\res{E3.mimo.lo}, \res{E3.mimo.hi}] for MiMo (\res{E3.mimo.p}) and \res{E3.gp.mean} [\res{E3.gp.lo}, \res{E3.gp.hi}] for the GP Reader (\res{E3.gp.p}). Every pooled interval lies within the 2~\eur{} effect of interest and well short of the E1 gap. Even the baseline's numerical Reader, given a model's evidence, gained on average at most \res{range.E3.gp.llm.hi}~\eur{} on any model's packets, a small fraction of the E1 gap. Better reading of the same evidence would therefore close little of the gap; the cause more likely lies earlier, in where the models start and which policies they try. As a consistency check, the GP Reader returned exactly the baseline's final policy on all 16 baseline packets. By source the picture is mixed and, with 16 packets per cell, descriptive only.

\subsection{Headroom, sensitivity and cost}
\label{sec:robust}
Three further analyses qualify the results above: how far better policies lie above the baseline, how much the E1 gap depends on two environment assumptions, and what a campaign costs.
\paragraph{Headroom.} On \res{ref.sites} random test sites, a numerical search for the best feasible policy found policies that beat the fixed reference by \res{ref.headroom}~\eur{} on average; two repeated searches differed by \res{ref.err.lo} and \res{ref.err.hi}~\eur{}, so the best-known values are lower bounds with that error. On these sites the baseline's final policies were \res{ref.baseline.vsfixed} relative to the fixed reference, most of the estimated headroom, while the models' were \res{ref.glm.vsfixed} (GLM), \res{ref.mimo.vsfixed} (MiMo) and \res{ref.deepseek.vsfixed} (DeepSeek).

\paragraph{Sensitivity.} The size of the E1 gap depends on two environment assumptions. With the side-wall heat exchange set to 4~W/m$^2$K instead of each site's drawn value (1--3) on \res{ueff.sites} random sites, the model-minus-baseline differences shrank to \res{ueff.4.deepseek} (DeepSeek), \res{ueff.4.glm} (GLM) and \res{ueff.4.mimo}~\eur{} (MiMo), against \res{ueff.main.deepseek}, \res{ueff.main.glm} and \res{ueff.main.mimo} on the same sites at the drawn value; the method ranking \res{ueff.4.rankchange} at 4~W/m$^2$K and was \res{ueff.0.rankchange} at 0. With a fruit dry-matter fraction of \setting{0.07} instead of \setting{0.06} the E1 differences shrank to \res{dm.0.07.deepseek} (DeepSeek), \res{dm.0.07.glm} (GLM) and \res{dm.0.07.mimo} (MiMo); at \setting{0.05} they grew to \res{dm.0.05.deepseek}, \res{dm.0.05.glm} and \res{dm.0.05.mimo}. The sign of E1 held in every setting except MiMo at the highest heat exchange, and E2 was insensitive to the dry-matter fraction.

\paragraph{Cost and failure modes.} A branched campaign cost on average USD \res{cost.deepseek} (DeepSeek, \res{calls.deepseek} calls), \res{cost.glm} (GLM, \res{calls.glm}) and \res{cost.mimo} (MiMo, \res{calls.mimo}). MiMo produced markedly more format errors (\res{formaterr.mimo} per Full branch against \res{formaterr.deepseek} and \res{formaterr.glm}); these are part of the method's result.

\section{Discussion and limitations}
\label{sec:discussion}
\paragraph{Where the agents fall short.} The three questions locate the models' weakness in the same place. They started below a standard grower setting and did not gain enough within the year to pass it (RQ1); they ran the four compartments as two synchronous batches and so never let in-season measurements change what they planted (RQ2); and the final reading step was not where they lost (RQ3). The bottleneck is how the agents used the facility as an instrument (where to start, what to try, when to plant), not how they summarised a year of outcomes.

\paragraph{Interpreting the baseline gap.} A simple trust-region search that starts from a standard grower setting and recommends only what it tried beat the models and reached most of the headroom found by an offline search. It brings domain knowledge the models lack, so the result shows how far general-purpose agents are from a domain-informed optimiser in this task, not that language models cannot do better; giving an agent the same starting point, or a scheduler that staggers experiments, are natural next tests on the same sites. Because the gap narrows when neighbouring spaces exchange more heat (Section~\ref{sec:robust}), the E1 numbers are not a general ranking of methods for greenhouse control.

\paragraph{Limitations.} SlowLab is a simulation study: the executor is not calibrated to a real greenhouse, its external check failed for temperature and humidity (Appendix~\ref{app:external}), and the sensor-noise model is a declared assumption. Results concern the tested facility, budget, action space and models. Three low-cost models were tested; stronger or reasoning-heavy models may behave differently, and GLM ran with mandatory capped reasoning. Only two environment sensitivities were run; no robustness claim is made for other assumptions. The baseline was redesigned after the pilot because the plain BO scored below the fixed reference; it was tuned only on development sites, but the redesign is a pilot-informed decision, and its radius was selected at the largest of the three preregistered candidates, so a wider search might make it stronger still. E3 drew each method's packets from different sites, so it bounds the effect of the reading step but cannot compare the quality of evidence across methods; a site-paired design would. The run was completed in two parts: the corrected baseline was rerun after the first part's results existed, and the corrected language-model branches were continued days after they started, so a provider-side change in between cannot be excluded (Appendix~\ref{app:original}).

\section{Conclusion}
\label{sec:conclusion}
SlowLab benchmarks agents whose experiments are slow, irreversible and bound to four compartments. On \res{sites} unseen sites, three low-cost language models stopped near a standard grower setting, clearly below a prior-informed local optimiser, and ran the facility as two synchronous batches that in-season measurements never changed. We release the benchmark and its locked protocol, and will release the run records, so that stronger agents can be measured on the same sites.

\paragraph{Code and data availability.} Code, locked configurations and analysis outputs: \url{https://github.com/dingningpei/slowlab-benchmark} (MIT licence). The campaign and evaluation records will be released under \ccby{}; the test-site and E3 seeds stay sealed (their salted commitments are public) so that the test sites remain unseen. Weather data: KNMI, \ccby{} \citep{knmi_cabauw,knmi_cabauw_rad}.

\bibliographystyle{plainnat}
\bibliography{refs}
\appendix
\section{Protocol, locks and audit}
\label{app:protocol}
\paragraph{Preregistration and locks.} Sample size, tests, model identities, prompts, code and configuration were frozen before the formal run: the run refused to start unless every locked file hash matched (167 files, six prompt hashes, method versions, model and provider identities, seed commitments and the run environment). The analysis code was locked while campaigns ran and before any formal score existed, and was not changed afterwards; the E3 Reader prompt and the remaining Phase~6 code were locked in the same way before any formal score was read. The test-site master seed and the E3 sample seed were committed with salted hashes in advance. The lock did not include the preregistration documents themselves; their hashes at the lock commit, which was pushed before the run, were recorded afterwards.

\paragraph{Execution.} The run used two rented 128-core servers. Evaluations and E3 packets repeated on the second server were bit-identical to the first's. The USD~30 API cap was enforced by provider-side limits summing to it, with pacing reserves in the runner.

\paragraph{Budget and locks.} The formal run used \res{campaigns} campaigns and \res{evaluations} evaluations. A development evaluation repeated on the first server was \res{parity.identical} to the earlier desktop run. API spend was USD \res{spend.original} in the first part (USD \res{spend.campaigns} for campaigns) and USD \res{spend.repair} in the second, USD \res{spend.total} in total, under a USD~30 cap. Locks: v1 before the run (code, configuration, prompts, model identities, seed commitments and environment), v2 the E1/E2 analysis code and v3 the Phase~6 code, both before any formal score existed, v4 the hashes of the preregistration documents at the v1 commit, and v6--v8 for the second part (v5 was superseded before use); every v6--v8 lock requires the rendered prompts to equal v1's and records the files changed since v1.

\paragraph{Integrity.} Every planned campaign and evaluation completed, so no data are missing. In the second part, \res{repair.failed} attempts that stopped on dropped provider connections were rerun from the same point and are kept, and one branch needed more than the preregistered two retries because of a defect in the model client, handled by a recorded decision (Appendix~\ref{app:original}). An audit confirmed one record per planned identity, that all \res{calls.total} model calls (\res{calls.repair} of them in the second part) were served by the planned model and provider with the planned settings, and that the booked spend equals the spend recomputed from per-call records.

\section{The original run and the correction}
\label{app:original}
\paragraph{The error.} The agent sessions receive the task description with keys in sorted order (canonical JSON), whereas the process predictor and the baseline's Gaussian-process kernel had been trained in the contract's field order, and the deployed code took policy fields in the order received. Validation had exercised only the offline path. On development data the misordering roughly doubled the predictor's error at day 90 (5.8 to 12.6~\eur{}) and lowered the kernel's cross-season ranking accuracy from 0.81 to 0.69 (decision record 2026-10-05). It affected only the predictor tool and the baseline: language-model Endpoint branches, pre-experiment recommendations and the shared day-0 designs never used them, and a scan of all \res{repair.audits} language-model audits found \res{repair.branches} Full branches that had received a predictor output (DeepSeek \res{repair.branches.deepseek}, GLM \res{repair.branches.glm}, MiMo \res{repair.branches.mimo}).

\paragraph{The correction.} The fix maps fields by name into the training order and adds a test comparing the deployed and offline paths. Under a new lock (prompts identical to lock v1): (i) the baseline's radius was re-selected on the same development sites by the same rule; (ii) all \res{repair.baseline} baseline campaigns were rerun with the same sites, years and seeds; (iii) each affected Full branch was replayed from its recorded conversation, with every outgoing message checked byte for byte against the original audit (\res{repair.replayed} calls), and continued live from the first predictor output (\res{repair.live} calls). Four branches stopped because the model client did not retry connections dropped after a request was sent, a defect that had not occurred in the first part. Two completed on their first retry. The other two failed again (one of them twice) until the client was fixed under a new lock with a recorded decision, and each then completed on its first retry. All \res{repair.failed} failed attempts are kept. (iv) the \res{repair.evaluations} changed recommendations were evaluated, the \res{repair.e3packets} E3 packets whose source records changed were rebuilt and re-read, and the boundary sensitivity was repeated for changed recommendations, while everything else was reused unchanged. (v) the locked analysis code was run once on the corrected data. The second part cost USD \res{spend.repair}. One host processor socket ran about 35 times slower than the other; jobs were then pinned to the fast socket, and evaluations repeated across sockets were \res{repair.socketparity}.

\paragraph{Original versus corrected results.} Table~\ref{tab:original} lists every primary and secondary test. The language models' own results barely changed; the baseline became stronger (Full mean \res{orig.mean.baseline.full} to \res{mean.baseline.full}~\eur{}), so all three E1 differences became significant, MiMo's gain over its pre-experiment recommendation lost significance, and the E3 Reader-swap losses of the original run, which were significant for all three models and largest on the baseline's packets, disappeared. Both analyses were computed on the same host: the Phase~6 analysis of the original data had first been computed on a different platform, where BCa bounds of samples with many ties differed (largest difference 0.46~\eur{} in one bound), without changing any test decision.

\begin{table}[h]
\centering\scriptsize
\caption{Original and corrected results (\eur{}; mean [BCa 95\% interval], $p$: Holm for E1/E2, BH over the joint secondary family otherwise).}
\label{tab:original}
\begin{tabular}{lll}
\toprule
Test & Original & Corrected \\
\midrule
E1 DeepSeek $-$ baseline & \res{orig.E1.deepseek.mean} [\res{orig.E1.deepseek.lo}, \res{orig.E1.deepseek.hi}], \res{orig.E1.deepseek.p} & \res{E1.deepseek.mean} [\res{E1.deepseek.lo}, \res{E1.deepseek.hi}], \res{E1.deepseek.p} \\
E1 GLM $-$ baseline & \res{orig.E1.glm.mean} [\res{orig.E1.glm.lo}, \res{orig.E1.glm.hi}], \res{orig.E1.glm.p} & \res{E1.glm.mean} [\res{E1.glm.lo}, \res{E1.glm.hi}], \res{E1.glm.p} \\
E1 MiMo $-$ baseline & \res{orig.E1.mimo.mean} [\res{orig.E1.mimo.lo}, \res{orig.E1.mimo.hi}], \res{orig.E1.mimo.p} & \res{E1.mimo.mean} [\res{E1.mimo.lo}, \res{E1.mimo.hi}], \res{E1.mimo.p} \\
E2 DeepSeek Full $-$ Endpoint & \res{orig.E2.deepseek.mean} [\res{orig.E2.deepseek.lo}, \res{orig.E2.deepseek.hi}], \res{orig.E2.deepseek.p} & \res{E2.deepseek.mean} [\res{E2.deepseek.lo}, \res{E2.deepseek.hi}], \res{E2.deepseek.p} \\
E2 GLM Full $-$ Endpoint & \res{orig.E2.glm.mean} [\res{orig.E2.glm.lo}, \res{orig.E2.glm.hi}], \res{orig.E2.glm.p} & \res{E2.glm.mean} [\res{E2.glm.lo}, \res{E2.glm.hi}], \res{E2.glm.p} \\
E2 MiMo Full $-$ Endpoint & \res{orig.E2.mimo.mean} [\res{orig.E2.mimo.lo}, \res{orig.E2.mimo.hi}], \res{orig.E2.mimo.p} & \res{E2.mimo.mean} [\res{E2.mimo.lo}, \res{E2.mimo.hi}], \res{E2.mimo.p} \\
E2 Baseline Full $-$ Endpoint & \res{orig.E2.baseline.mean} [\res{orig.E2.baseline.lo}, \res{orig.E2.baseline.hi}], \res{orig.E2.baseline.p} & \res{E2.baseline.mean} [\res{E2.baseline.lo}, \res{E2.baseline.hi}], \res{E2.baseline.p} \\
\midrule
DeepSeek final $-$ initial & \res{orig.learn.deepseek.mean} [\res{orig.learn.deepseek.lo}, \res{orig.learn.deepseek.hi}], \res{orig.learn.deepseek.p} & \res{learn.deepseek.mean} [\res{learn.deepseek.lo}, \res{learn.deepseek.hi}], \res{learn.deepseek.p} \\
GLM final $-$ initial & \res{orig.learn.glm.mean} [\res{orig.learn.glm.lo}, \res{orig.learn.glm.hi}], \res{orig.learn.glm.p} & \res{learn.glm.mean} [\res{learn.glm.lo}, \res{learn.glm.hi}], \res{learn.glm.p} \\
MiMo final $-$ initial & \res{orig.learn.mimo.mean} [\res{orig.learn.mimo.lo}, \res{orig.learn.mimo.hi}], \res{orig.learn.mimo.p} & \res{learn.mimo.mean} [\res{learn.mimo.lo}, \res{learn.mimo.hi}], \res{learn.mimo.p} \\
DeepSeek $-$ fixed reference & \res{orig.vsfixed.deepseek.mean} [\res{orig.vsfixed.deepseek.lo}, \res{orig.vsfixed.deepseek.hi}], \res{orig.vsfixed.deepseek.p} & \res{vsfixed.deepseek.mean} [\res{vsfixed.deepseek.lo}, \res{vsfixed.deepseek.hi}], \res{vsfixed.deepseek.p} \\
GLM $-$ fixed reference & \res{orig.vsfixed.glm.mean} [\res{orig.vsfixed.glm.lo}, \res{orig.vsfixed.glm.hi}], \res{orig.vsfixed.glm.p} & \res{vsfixed.glm.mean} [\res{vsfixed.glm.lo}, \res{vsfixed.glm.hi}], \res{vsfixed.glm.p} \\
MiMo $-$ fixed reference & \res{orig.vsfixed.mimo.mean} [\res{orig.vsfixed.mimo.lo}, \res{orig.vsfixed.mimo.hi}], \res{orig.vsfixed.mimo.p} & \res{vsfixed.mimo.mean} [\res{vsfixed.mimo.lo}, \res{vsfixed.mimo.hi}], \res{vsfixed.mimo.p} \\
Baseline $-$ fixed reference & \res{orig.vsfixed.baseline.mean} [\res{orig.vsfixed.baseline.lo}, \res{orig.vsfixed.baseline.hi}], \res{orig.vsfixed.baseline.p} & \res{vsfixed.baseline.mean} [\res{vsfixed.baseline.lo}, \res{vsfixed.baseline.hi}], \res{vsfixed.baseline.p} \\
Feedback gain DeepSeek $-$ baseline & \res{orig.gaindiff.deepseek.baseline.mean} [\res{orig.gaindiff.deepseek.baseline.lo}, \res{orig.gaindiff.deepseek.baseline.hi}], \res{orig.gaindiff.deepseek.baseline.p} & \res{gaindiff.deepseek.baseline.mean} [\res{gaindiff.deepseek.baseline.lo}, \res{gaindiff.deepseek.baseline.hi}], \res{gaindiff.deepseek.baseline.p} \\
Feedback gain DeepSeek $-$ GLM & \res{orig.gaindiff.deepseek.glm.mean} [\res{orig.gaindiff.deepseek.glm.lo}, \res{orig.gaindiff.deepseek.glm.hi}], \res{orig.gaindiff.deepseek.glm.p} & \res{gaindiff.deepseek.glm.mean} [\res{gaindiff.deepseek.glm.lo}, \res{gaindiff.deepseek.glm.hi}], \res{gaindiff.deepseek.glm.p} \\
Feedback gain DeepSeek $-$ MiMo & \res{orig.gaindiff.deepseek.mimo.mean} [\res{orig.gaindiff.deepseek.mimo.lo}, \res{orig.gaindiff.deepseek.mimo.hi}], \res{orig.gaindiff.deepseek.mimo.p} & \res{gaindiff.deepseek.mimo.mean} [\res{gaindiff.deepseek.mimo.lo}, \res{gaindiff.deepseek.mimo.hi}], \res{gaindiff.deepseek.mimo.p} \\
Feedback gain GLM $-$ baseline & \res{orig.gaindiff.glm.baseline.mean} [\res{orig.gaindiff.glm.baseline.lo}, \res{orig.gaindiff.glm.baseline.hi}], \res{orig.gaindiff.glm.baseline.p} & \res{gaindiff.glm.baseline.mean} [\res{gaindiff.glm.baseline.lo}, \res{gaindiff.glm.baseline.hi}], \res{gaindiff.glm.baseline.p} \\
Feedback gain GLM $-$ MiMo & \res{orig.gaindiff.glm.mimo.mean} [\res{orig.gaindiff.glm.mimo.lo}, \res{orig.gaindiff.glm.mimo.hi}], \res{orig.gaindiff.glm.mimo.p} & \res{gaindiff.glm.mimo.mean} [\res{gaindiff.glm.mimo.lo}, \res{gaindiff.glm.mimo.hi}], \res{gaindiff.glm.mimo.p} \\
Feedback gain MiMo $-$ baseline & \res{orig.gaindiff.mimo.baseline.mean} [\res{orig.gaindiff.mimo.baseline.lo}, \res{orig.gaindiff.mimo.baseline.hi}], \res{orig.gaindiff.mimo.baseline.p} & \res{gaindiff.mimo.baseline.mean} [\res{gaindiff.mimo.baseline.lo}, \res{gaindiff.mimo.baseline.hi}], \res{gaindiff.mimo.baseline.p} \\
E3 GP Reader swap gain & \res{orig.E3.gp.mean} [\res{orig.E3.gp.lo}, \res{orig.E3.gp.hi}], \res{orig.E3.gp.p} & \res{E3.gp.mean} [\res{E3.gp.lo}, \res{E3.gp.hi}], \res{E3.gp.p} \\
E3 DeepSeek swap gain & \res{orig.E3.deepseek.mean} [\res{orig.E3.deepseek.lo}, \res{orig.E3.deepseek.hi}], \res{orig.E3.deepseek.p} & \res{E3.deepseek.mean} [\res{E3.deepseek.lo}, \res{E3.deepseek.hi}], \res{E3.deepseek.p} \\
E3 GLM swap gain & \res{orig.E3.glm.mean} [\res{orig.E3.glm.lo}, \res{orig.E3.glm.hi}], \res{orig.E3.glm.p} & \res{E3.glm.mean} [\res{E3.glm.lo}, \res{E3.glm.hi}], \res{E3.glm.p} \\
E3 MiMo swap gain & \res{orig.E3.mimo.mean} [\res{orig.E3.mimo.lo}, \res{orig.E3.mimo.hi}], \res{orig.E3.mimo.p} & \res{E3.mimo.mean} [\res{E3.mimo.lo}, \res{E3.mimo.hi}], \res{E3.mimo.p} \\
\bottomrule
\end{tabular}
\end{table}

\section{Baseline redesign}
\label{app:baseline}
The preregistered plan first named a plain asynchronous GP/BO as the only numerical comparator. In the eight-site pilot it scored \res{pilot.bo.vsfixed}~\eur{} relative to the fixed reference, which uses no experiments at all, and its two repeats on the same site differed with an SD of \res{pilot.bo.repeatsd}~\eur{} (language models: \res{pilot.llm.repeatsd.lo}--\res{pilot.llm.repeatsd.hi}). A comparator below the zero-experiment reference is not a meaningful baseline, so it was replaced before any test-site campaign was run; the decision referred only to the fixed reference. Its pilot and development results are reported here and it was not run on test sites.

The replacement differs in three ways, all set on development sites only. (i) It starts from the fixed reference and explores within a trust region around the best tried policy. (ii) Its Gaussian-process hyperparameters are frozen from development data (\res{kernel.groups} site-years, \res{kernel.crops} crops with random policies) instead of being refitted from two to six crops in each campaign. A kernel that separates a season effect common to all policies from the policy effect was chosen over a product kernel by a rule fixed before the fit: it ranked second-season crops from first-season data with accuracy \res{kernel.cross.interaction} against \res{kernel.cross.product}, and its held-out error on campaign-sized data was \res{kernel.rmse.interaction}~\eur{} against \res{kernel.rmse.refit} for per-campaign refitting and \res{kernel.rmse.mean} for the mean of the observed crops. (iii) Only tried policies are recommended. The initial trust-region radius was chosen among 15, 25 and 35\% of each range by mean development score (\res{radius.15.mean}, \res{radius.25.mean} and \res{radius.35.mean}~\eur{}; fixed reference \res{radius.fixed}; plain BO \res{radius.v3.mean}); the selected \res{radius.selected}\% is the largest candidate, and the frozen rule allowed no further candidates. These are the scores of the corrected agent; with the field-order error (Appendix~\ref{app:original}) the same rule had selected \res{orig.radius.selected}\% (\res{orig.radius.15.mean}, \res{orig.radius.25.mean} and \res{orig.radius.35.mean}~\eur{}). On the pilot sites its repeat SD was \res{pilot.baseline.repeatsd}~\eur{}.

\section{External validity check}
\label{app:external}
Before the study we tested whether GreenLight reproduces observed climate when driven by the actions of the 2019 Autonomous Greenhouse Challenge teams, in a one-shot temporal holdout with thresholds fixed in advance. It failed: pooled RMSE was \res{agc.temp.rmse}\,$^\circ$C for air temperature (threshold \res{agc.temp.limit}) and \res{agc.rh.rmse} percentage points for humidity (threshold \res{agc.rh.limit}); CO$_2$ passed (\res{agc.co2.rmse}~ppm, threshold \res{agc.co2.limit}). SlowLab is therefore a simulated benchmark; its numbers are not predictions for a real greenhouse.

\section{Agent interface details}
\label{app:interface}
\begin{figure}[h]
\centering
\includegraphics[width=\linewidth]{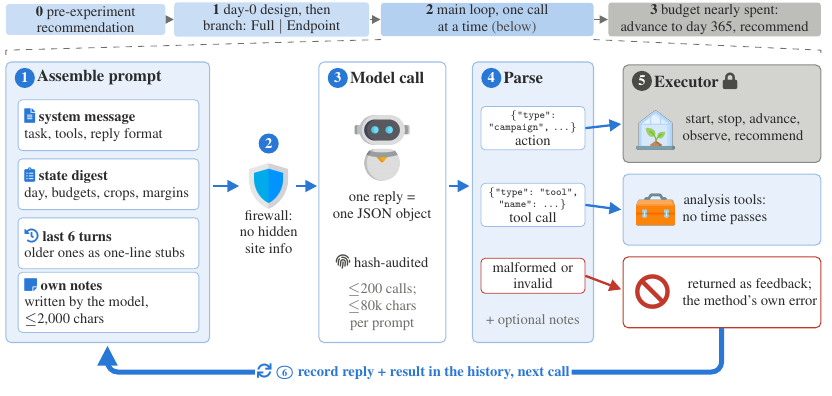}
\caption{The language-model harness. Top: the stages of a campaign. Bottom: one call of the main loop. The prompt is assembled from the system message, a digest of the state, the six most recent turns and the model's own notes (1); it passes the information firewall (2); the model returns one JSON object, and every message is hash-audited (3); the reply is parsed (4) and sent to the executor or the analysis tools, or returned as an error (5); the reply and its result join the history for the next call (6). Icons: Freepik, flaticon.com.}
\label{fig:harness}
\end{figure}
\paragraph{Messages.} The system message states the role (``You run a greenhouse experimentation campaign as the experimenter. Plan experiments, read the evidence your feedback condition allows, and recommend one management policy at the final day.''), then the public task description, the tool catalogue and the reply format, all as canonical JSON. Every later call ends with one user message holding the state digest, the latest result and the instruction. The reply format asks for exactly one JSON object: either \texttt{\{"type": "campaign", "action": \{...\}, "reason": "..."\}} for one of the five actions, or \texttt{\{"type": "tool", "name": "...", "args": \{...\}\}} for one analysis tool; any reply may carry \texttt{"notes"} to replace the notebook. The pre-experiment recommendation uses a separate grower prompt with the task only. The full prompts are hashed in the lock and released with the code.
\paragraph{Tools.} \texttt{runs} (every crop started, with released totals and margins); \texttt{series\_summary} (daily count, mean, minimum and maximum of a channel already read); \texttt{fit\_outcome\_model} (a Gaussian process of crop margin on policy and planting date, fitted to completed crops); \texttt{predict} (margins of up to 32 policies from that fit); \texttt{predict\_crop\_outcome} (the process predictor: final harvest, resources and margin of a running crop from its readings so far); \texttt{space\_filling\_candidates} (up to 64 feasible policies spread over the allowed ranges). Tools analyse only data the agent has already received; they do not advance time.
\paragraph{Context.} At most 80{,}000 characters per prompt and $5\times10^6$ per campaign; the six most recent results are shown in full and older ones as one-line stubs; reasons are truncated to 400 characters; at most 200 model calls. When the remaining characters fall below one prompt plus a reserve, or the call budget is nearly spent, the harness advances to day 365 and accepts only a recommendation.

\section{Example campaign}
\label{app:trace}
\begin{figure}[h]
\centering
\includegraphics[width=0.85\linewidth]{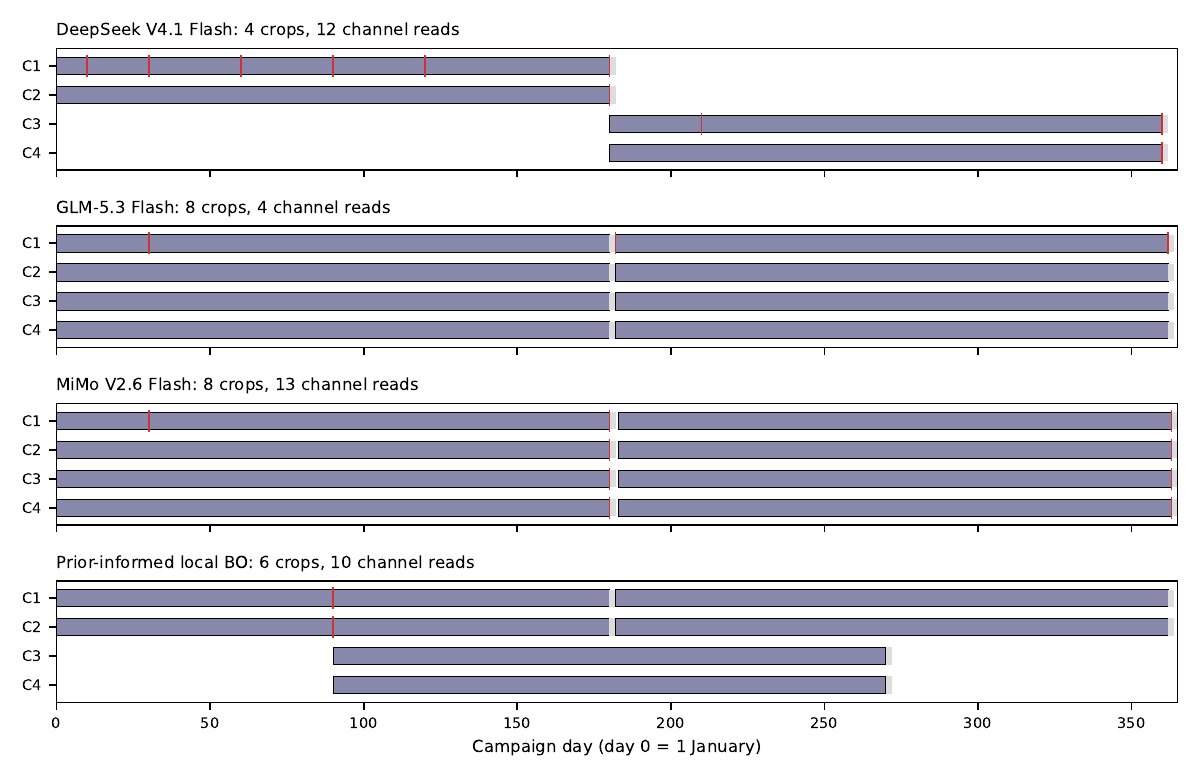}
\caption{Four Full-feedback campaigns on test site 0 (selection fixed before plotting: first repeat of each model, first seed of the baseline). Bars are crops in the four compartments C1--C4 (grey: cleaning), red ticks are channel reads. The models plant their second wave only when the first closes; the baseline staggers two compartments at mid-season.}
\label{fig:timelines}
\end{figure}
Table~\ref{tab:trace} lists the Full branch of the campaign in Figure~\ref{fig:teaser} (test site 0, first repeat, DeepSeek) with the model's stated reasons, quoted verbatim. Where a row covers several steps, the reason of the first is shown. The model calls its own day-0 policy its ``baseline''; this is not the BO baseline.
\begin{table}[h]
\centering\scriptsize
\caption{The Full branch of the campaign in Figure~\ref{fig:teaser}: day, action and the model's stated reason (verbatim).}
\label{tab:trace}
\begin{tabular}{@{}r p{0.2\linewidth} p{0.66\linewidth}@{}}
\toprule
Day & Action & Stated reason \\
\midrule
0 & advance to day 10 & Let the crops establish before reading climate and canopy data. \\
10 & observe (2 channels) & Check climate tracking in compartment 0 so far. \\
30--120 & observe, advance (9 steps) & Check canopy growth progress toward maturity. \\
150 & advance to day 180 & Let the baseline crops finish their 180-day cycle so totals are released. \\
180 & runs, observe (4 steps) & Get accrued resources and contribution margin for the two completed baseline crops. \\
180 & start (C3, C4; 2 steps) & Now let me explore a more aggressive policy in compartments 2 and 3, and start planning. \\
180--240 & advance, observe (4 steps) & Let the high- and low-input crops establish before comparing. \\
360 & fit\_outcome\_model & High-input policy lost money ($-$15.6) due to light/heat cost; low-input made 8.8; baseline made $\sim$20--22. Baseline looks best so far. Let me fit the outcome model with the completed crops. \\
365 & recommend & Baseline policy delivered the best margins (22.45 and 20.49 EUR/m2), far above low-input (8.79) and high-input ($-$15.58, where extra light/heat/CO2 cost outweighed yield gains). Recommend the baseline policy. \\
\bottomrule
\end{tabular}
\end{table}

\section{Environment specification}
\label{app:env}
Crops are stopped for safety if indoor air stays below 5\,$^\circ$C or above 40\,$^\circ$C for an hour or if they fail physiologically.
\begin{table}[h]
\centering\small
\caption{The SlowLab environment (task contract v8, site distribution v1). ``Assumed'' marks declared assumptions without a calibration source; ``literature'' marks ranges constrained by published values.}
\label{tab:task}
\begin{tabular}{@{}p{0.2\linewidth}p{0.48\linewidth}p{0.24\linewidth}@{}}
\toprule
Element & Specification & Basis \\
\midrule
Facility & 4 independent compartments of 96~m$^2$; GreenLight \texttt{fa502ed}; Cabauw weather & simulator; audited weather \\
Calendar & 365-day campaign; crop 180 d; cleaning 2 d; last start day 183; control step 300~s & benchmark design \\
Policy & day 18--26\,$^\circ$C; night 15--22\,$^\circ$C ($\le$ day); CO$_2$ 400--1200~ppm; light 0--16~h; vent offset 2--6\,$^\circ$C; RH limit 70--90\% & design bounds \\
Actions & start, stop (crop discarded), advance, observe, recommend; analysis tools incl.\ process predictor & benchmark design \\
Observations & air temperature, RH, CO$_2$, canopy LAI proxy; cumulative harvest, heat, light, CO$_2$ dosing; every 300~s & sensor noise v0 (assumed) \\
Feedback & Full: all records at any time; Endpoint: status while running, totals at close & experimental factor (E2) \\
Budget & 16 starts; 32 decision calls; 128 tool calls; $5\times10^6$ prompt characters (models) & benchmark design \\
Score & mean margin of two crops (1 Jan, 2 Jul) in 3 new weather years, at public site prices & private evaluator \\
Site: cultivar & temperature-window shift $\pm1.5$\,$^\circ$C; photosynthesis and fruit growth $\times$0.85--1.15; earliness $\times$0.9--1.1; max LAI 2.5--3.5 & assumed \\
Site: facility & cover transmission 0.75--0.88; leakage $3\cdot10^{-5}$--$2\cdot10^{-4}$; LED efficacy 0.45--0.6 & literature \\
 & ventilation $\times$0.8--1.2; side-wall exchange 1--3~W/m$^2$K; deep soil 8--20\,$^\circ$C & assumed \\
Site: other & prices (public); unobservable per-compartment differences; campaign and evaluation years & declared \\
\bottomrule
\end{tabular}
\end{table}

\section{Environment card and broader impact}
\label{app:card}
\paragraph{Environment card.} \emph{Purpose:} evaluating how agents plan, monitor and read slow, irreversible experiments in a shared facility. \emph{Composition:} a site distribution (cultivar, facility, price and per-compartment parameters), audited Cabauw weather years (KNMI, \ccby{}), the GreenLight model at a pinned commit, the task contract, the agent interface and tools, and a private evaluator. \emph{Splits:} development sites and years for tuning, eight pilot sites for power, \res{sites} test sites drawn from a committed master seed that stays sealed. \emph{Released records:} public campaign transcripts, model calls and replies with hashes, evaluation records and history packets (\ccby{}); private site parameters, seeds and simulation traces are not released. \emph{Known limitations:} not calibrated to a real greenhouse (Appendix~\ref{app:external}); declared assumptions are marked in Table~\ref{tab:task}. \emph{Maintenance:} the code repository records every change, and every frozen record is hash-locked.
\paragraph{Broader impact.} SlowLab measures agents in a simulator; it involves no human subjects or personal data, and its results are not advice for real greenhouse operation. A benchmark of autonomous experimentation could encourage deploying agents in real facilities before they are reliable; our results argue for caution, since the tested agents left most of the available improvement unused and did not act on in-season information. Sealing the test seeds keeps the test sites useful for future agents.

\end{document}